\documentclass{article}

\usepackage{PRIMEarxiv}

\usepackage[utf8]{inputenc} 
\usepackage[T1]{fontenc}    
\usepackage{hyperref}       
\usepackage{url}            
\usepackage{booktabs}       
\usepackage{amsfonts}       
\usepackage{nicefrac}       
\usepackage{microtype}      
\usepackage{lipsum}
\usepackage{fancyhdr}       
\usepackage{graphicx}       
\graphicspath{{media/}}     
\usepackage{graphicx}%
\usepackage{multirow}%
\usepackage{amsmath,amssymb,amsfonts}%
\usepackage{amsthm}%
\usepackage{mathrsfs}%
\usepackage{xcolor}%
\usepackage{multirow}
\usepackage{makecell} 
\usepackage{tabularx} 
\usepackage{colortbl}
\usepackage{enumitem}
\title{HAVE: Head-Adaptive Gating and ValuE Calibration for Hallucination Mitigation in Large Language Models}

\author{
   Xin Tong \\
  School of Information and Cyber Security \\
  People’s Public Security University of China \\
  Beijing\\
  \texttt{tongxindotnet@outlook.com} \\
  \And
Zhi Lin \\
  School of Safety Science \\
  Tsinghua University \\
  Beijing\\
  \texttt{linz20@mails.tsinghua.edu.cn} \\
  \And
Jingya Wang \\
  School of Information and Cyber Security \\
  People’s Public Security University of China \\
  Beijing\\
  \texttt{wangjingya@ppsuc.edu.cn} \\
  \And
Bo Jin* \\
  The Third Research Institute of the Ministry of Public Security of China \\
  Shanghai\\
  \texttt{jinbo@gass.cn} \\
}

\begin{document}
\maketitle

\begin{abstract}
Large Language Models (LLMs) often produce hallucinations in retrieval-augmented or long-context generation, even when relevant evidence is present. This stems from two issues: head importance is treated as input-agnostic, and raw attention weights poorly reflect each token’s true contribution. We present HAVE (Head-Adaptive Gating and ValuE Calibration), a parameter-free decoding framework that directly addresses both challenges. HAVE introduces head-adaptive gating, which performs instance-level soft reweighing of attention heads, and value calibration, which augments attention with the magnitude of value vectors to approximate write-back contribution. Together, these modules construct token-level evidence aligned with model updates and fuse it with the LM distribution through a lightweight uncertainty-scaled policy. HAVE requires no finetuning and operates in a single forward pass, making it efficient and broadly applicable. Experiments across multiple QA benchmarks and LLM families demonstrate that HAVE consistently reduces hallucinations and outperforms strong baselines, including DAGCD, with modest overhead. The framework is transparent, reproducible, and readily integrates with off-the-shelf LLMs, advancing trustworthy generation in real-world settings.

\end{abstract}


\section{Introduction}
Large language models (LLMs) are increasingly deployed in retrieval-augmented \cite{lewis2020retrieval} and long-context scenarios where correct answers should be directly grounded in supplied evidence. Yet hallucinations persist: even when attention patterns highlight relevant spans, the generated output can drift away from the facts. This gap arises because attention does not perfectly reflect how internal signals shape next-token predictions, leaving faithful decoding an open challenge.

Prior work \cite{huang2025dynamic} has explored leveraging attention maps to guide decoding, but two fundamental issues remain. First, head importance is typically treated as static, despite clear variation across domains and inputs. Relying on fixed head weights underutilizes context-sensitive heads and amplifies noisy ones. Second, raw attention is a poor proxy for contribution to the residual stream update that ultimately determines token probabilities. Tokens that are frequently “read” may exert little influence, while “sink” tokens such as special or whitespace symbols can dominate. These limitations hinder the reliability of attention-based evidence.

We propose HAVE (\textbf{H}ead-\textbf{A}daptive Gating and \textbf{V}alu\textbf{E} Calibration), a single-step, parameter-free framework for faithful decoding.
HAVE is efficient: it requires only the standard forward pass of the base model, introduces no new parameters, and is robust to grouped-query attention and sliding-window key–value caches. Moreover, its internal signals—dynamic head weights and calibrated token scores—are interpretable and can be visualized or ablated, providing diagnostic value beyond performance gains.

Specifically, this work makes the following contributions:

\begin{itemize}
\item We introduce a new perspective on decoding-time evidence, showing that raw attention is insufficient and proposing to calibrate it with value signals that directly approximate contribution to the residual stream.

\item We develop HAVE, a parameter-free framework with two novel modules: (i) Head-Adaptive Gating, which performs instance-level soft reweighting of attention heads, and (ii) Value Calibration, which debiases sinks and augments attention with value norms to produce contribution-aware token evidence.

\item We provide a general and efficient implementation that operates in a single forward pass and is compatible with modern architectures (grouped-query attention, sliding-window caches), while exposing interpretable internal signals (dynamic head weights, calibrated token scores).
\end{itemize}

Through comprehensive experiments on QA benchmarks and across LLM families, HAVE achieves consistent improvements over strong attention-guided baselines, establishing new state-of-the-art in several settings with modest overhead.

By combining interpretability with effectiveness, HAVE offers a practical step toward trustworthy LLM generation in retrieval-augmented and long-context settings.

\section{Related Work}
Across the entire pipeline—from user input to model output—factors such as input data, model training, and inference strategies all play critical roles in shaping factual accuracy. To address these challenges, prior research has explored three major categories of enhancement methods: training-based improvements, input-based strategies, and output-level optimizations.

Training-based enhancements have emerged as a prominent area of focus in improving factual accuracy. These approaches incorporate augmentation mechanisms during pre-training \cite{li2023textbooks}, fine-tuning \cite{hu2024mitigating}, and preference alignment \cite{sun2024aligning}. By adjusting the model’s memory weights, they aim to maintain generative fluency while simultaneously ensuring greater factual consistency.

Another line of research leverages the emergent properties of LLMs, including in-context learning, instruction following, and chain-of-thought reasoning. Prompt engineering, in particular, has been shown to improve factual robustness. For example, carefully designed prompts can encourage models to assess their own confidence and perform self-correction \cite{li2024confidence}. Similarly, in-context demonstrations can provide models with updated examples, helping them maintain critical facts while filtering out irrelevant information \cite{zheng2023can}. Such strategies effectively enable “temporary” knowledge updates without modifying model parameters. However, their reliability in handling complex, dynamically evolving knowledge domains or multi-step reasoning tasks still requires further validation. Complementary to these methods, the Retrieval-Augmented Generation (RAG) framework \cite{lewis2020retrieval} strengthens factuality by integrating external knowledge bases through indexing, retrieval, and generation.

Finally, output optimization strategies aim to refine factuality during or after the decoding process. One class of methods optimizes the decoding process of LLMs, enhancing the factual consistency of their outputs without modifying the model parameters. CAD~\cite{shi2024trusting} computes both a “context-aware” output distribution and a “context-free” output distribution. By contrasting the two, it reorders the probabilities of the next token, thereby reinforcing words that “only become more likely after reading the context.” COIECD~\cite{yuan2024discerning} first applies an information-theoretic approach to classify tokens into or out of a constraint set, depending on whether they conflict with parametric knowledge. It then applies different contrastive decoding strategies to conflict and non-conflict tokens.DAGCD~\cite{huang2025dynamic} leverages attention signals together with uncertainty estimates to dynamically amplify the contribution of tokens that are genuinely utilized from the context, thereby optimizing the decoding process. Another approach applies multi-stage correction or iterative editing \cite{krishna2024genaudit, chen2023converge}, thereby compensating for factual shortcomings in the generation phase.

Together, these lines of research highlight the growing recognition that improving factuality in LLMs requires a multifaceted approach, spanning training, input, and output dimensions.

\section{Method}
\label{sec:method}

\subsection{Preliminaries}
\label{sec:prelim}

At decoding step $t$, a frozen autoregressive language model over vocabulary $V$ produces a distribution
\begin{equation}
\label{eq:decoding}
P_t = \mathbb{softmax}(z_t) \in \mathbb{R}^{|V|},
\end{equation}
conditioned on a visible context sequence $X=\{x_j\}_{j=1}^{|X|}$ determined by the model’s key--value (KV) cache.

We denote by $\mathcal{C}\subseteq\{1,\dots,|X|\}$ the set of context indices visible to step $t$.
For each layer $\ell \in \{1,\dots,L\}$ and head $h$, let
\begin{itemize}
    \item $a_{\ell,h}(j)\in[0,1]$ be the last-row attention weight from step $t$ to position $j\in\mathcal{C}$, and
    \item $V_{\ell,h}(j)\in\mathbb{R}^{d}$ the corresponding value vector retrieved from the KV cache.
\end{itemize}

Finally, let $\mathcal{R}_t \subseteq V$ denote the Top-$R$ support of $P_t$, i.e., the set of $R$ highest-probability vocabulary items at step $t$.

\subsection{Framework Overview}
\label{sec:overview}

We factor faithful decoding into two components: 
(i) \emph{evidence construction} from internal activations, and 
(ii) a \emph{fusion policy} that integrates this evidence with the model distribution. 
Formally, at each decoding step $t$,
\begin{equation}
\label{eq:phi}
U_t \;=\; \Phi\!\big(X,\{a_{\ell,h}\},\{V_{\ell,h}\}\big)\in\mathbb{R}^{|V|},
\end{equation}
\begin{equation}
\label{eq:psi}
S_t \;=\; \Psi\!\big(P_t, U_t\big)\in\mathbb{R}^{|V|}.
\end{equation}
Here $\Phi$ maps attention and value activations to a vocabulary-aligned \emph{utilization} distribution $U_t$, and 
$\Psi$ combines $U_t$ with the LM distribution $P_t$ under a lightweight gating policy.

Our design is guided by three principles:
\begin{enumerate}[label=(\roman*)]
\item \textbf{Single-pass:} $\Phi$ and $\Psi$ rely only on activations from the current forward step, requiring no finetuning or additional passes.
\item \textbf{Support preservation:} outside the model’s Top-$R$ candidates $\mathcal{R}_t$, the distribution remains unchanged:
\begin{equation}
\label{eq:support}
S_t(v) \;=\; P_t(v), \qquad v \notin \mathcal{R}_t.
\end{equation}
\item \textbf{Monotonicity on support:} within $\mathcal{R}_t$, stronger evidence cannot reduce a candidate’s score:
\begin{equation}
\label{eq:monotone}
U_t(v_1)\ge U_t(v_2), \; v_1,v_2\in \mathcal{R}_t 
\;\;\Rightarrow\;\; S_t(v_1)\ge S_t(v_2).
\end{equation}
\end{enumerate}

\begin{figure}
  \centering
  \includegraphics[scale=0.12]{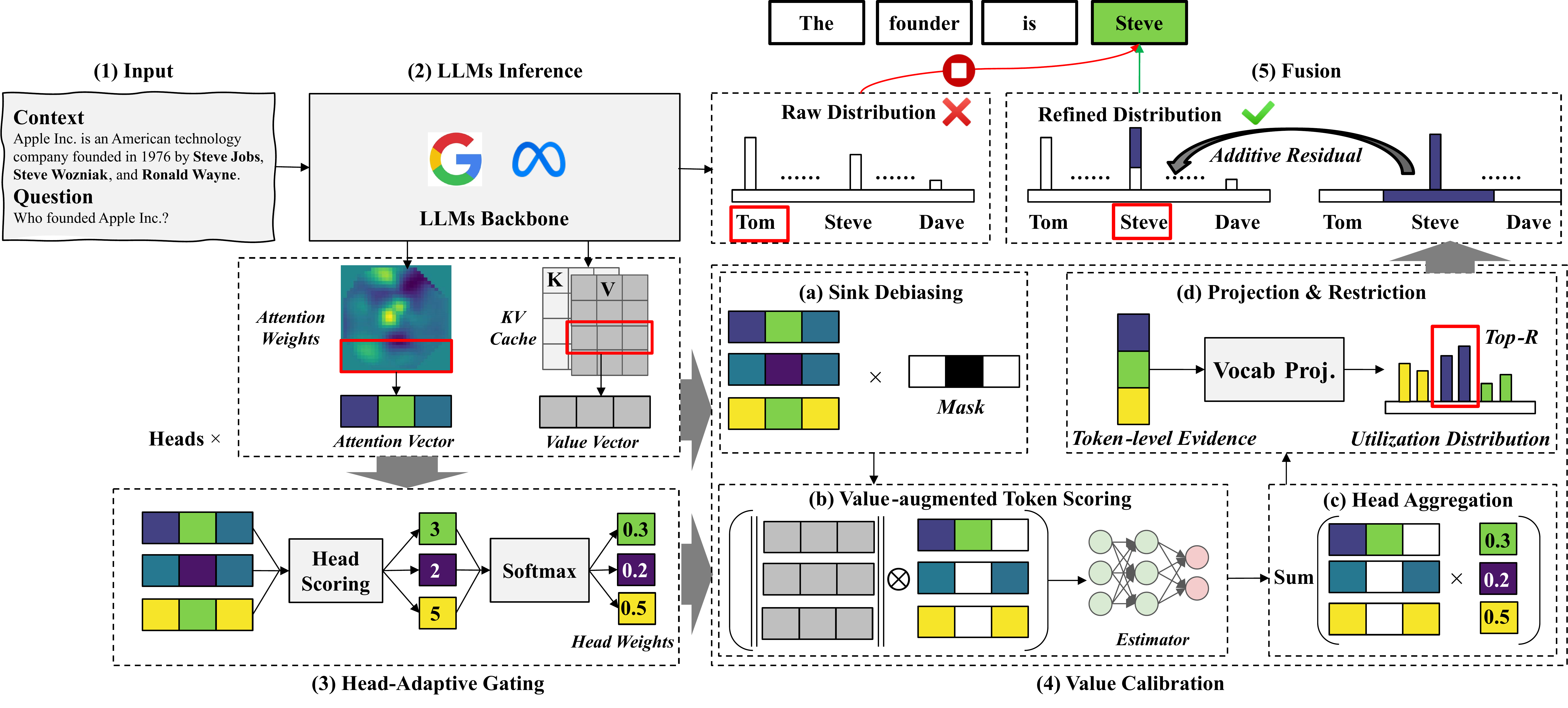}
  \caption{Overall Framework of HAVE.}\label{fig:framework}
\end{figure}

\paragraph{Our instantiation: \textbf{HAVE}.}
We realize $\Phi$ through two modules:
\emph{Head-Adaptive Gating} dynamically reweights attention heads at the instance level to obtain $w_{\ell,h}$, and 
\emph{Value Calibration} debiases sink tokens and augments attention with value norms ($\text{Attn}\times\|V\|$) to yield contribution-aware token evidence. 
The resulting distribution $U_t$ is restricted to $\mathcal{R}_t$, and the fusion policy $\Psi$ combines $U_t$ with $P_t$ via an uncertainty-scaled additive rule (Eq.~\ref{eq:fusion}). 
Fig~\ref{fig:framework} illustrates the framework of HAVE.

\subsection{Head-Adaptive Gating}
\label{sec:hag}
The first module of \textbf{HAVE} adaptively reweights attention heads at the \emph{instance level}.
Unlike static or hard Top-$K$ selection, \emph{Head-Adaptive Gating (HAG)} assigns dynamic yet strictly positive weights $w_{\ell,h}$, so that context-sensitive heads contribute more while all heads retain a small nonzero role for stability.

\paragraph{Instance-wise scoring.}
For each $(\ell,h)$ we compute a context-sensitivity score $s_{\ell,h}>0$ from the last-row attention over the visible context $\mathcal{C}$ (deduplicated but before sink masking):
\begin{equation}
\label{eq:head-mass}
s_{\ell,h} \;=\; \sum_{j\in\mathcal{C}} \bar a_{\ell,h}(j)\,\omega(j),
\qquad
\omega(j)\;=\;\frac{1}{|\{k\in\mathcal{C}:\mathrm{tok}(x_k)=\mathrm{tok}(x_j)\}|}.
\end{equation}

\paragraph{Soft gating.}
Instance weights are obtained by a normalized softmax over the scores $s_{\ell,h}$:
\begin{equation}
\tilde{w}^{(\mathrm{inst})}_{\ell,h}
= \frac{\exp(\log(s_{\ell,h}+\epsilon))}{\sum_{\ell',h'} \exp(\log(s_{\ell',h'}+\epsilon))},
\end{equation}
and combined with optional base priors $w^{(\mathrm{base})}_{\ell,h}$ under a small floor $\eta>0$:
\begin{equation}
w_{\ell,h}
\propto \max\!\Big\{ w^{(\mathrm{base})}_{\ell,h}\cdot \tilde{w}^{(\mathrm{inst})}_{\ell,h},\; \eta \Big\},
\qquad \sum_{\ell,h} w_{\ell,h}=1.
\end{equation}
This \emph{soft gating} preserves strictly positive weights for all heads, 
ensuring stability across instances without resorting to hard Top-$K$ selection.

\subsection{Value Calibration}
\label{sec:vc}
The second module of \textbf{HAVE} augments attention with value information to construct a token-level evidence signal that better approximates the write-back update at step $t$. 
While raw attention can overemphasize sink tokens or misalign with residual contributions, \emph{Value Calibration (VC)} suppresses such artifacts and produces contribution-aware evidence.

\paragraph{Sink correction.}
We first suppress sink tokens (special or whitespace) via a binary mask $M(j)$ and renormalize:
\begin{equation}
\label{eq:sink}
\tilde{a}_{\ell,h}(j)
= \frac{a_{\ell,h}(j)\,M(j)}{\sum_{k\in\mathcal{C}} a_{\ell,h}(k)\,M(k)+\varepsilon},
\qquad j\in\mathcal{C}.
\end{equation}

\paragraph{Value-augmented evidence.}
We weight attention by the magnitude of the value vector, approximating token contribution to the residual update:
\begin{equation}
\label{eq:attnV}
r_{\ell,h}(j)
= \tilde{a}_{\ell,h}(j)\,\|V_{\ell,h}(j)\|_{2},
\end{equation}
followed by normalization within each head:
\begin{equation}
\label{eq:head-norm}
\hat{r}_{\ell,h}(j)
= \frac{r_{\ell,h}(j)}{\sum_{k\in\mathcal{C}} r_{\ell,h}(k)+\varepsilon}.
\end{equation}

\paragraph{Refinement with lightweight estimator.}
To further calibrate token-level evidence, VC incorporates a lightweight estimator~\cite{huang2025dynamic} that consumes per-token feature vectors enriched with head-level channels and outputs a multiplicative mask $m(j)\in[0,1]$ for each token:
\begin{equation}
\label{eq:lr-mask}
m(j)=\sigma\!\big(w^{\top}f(j)+b\big),\qquad
U^{\mathrm{ctx}}_{t}(j)= m(j)\cdot \sum_{\ell,h} w_{\ell,h}\,\hat{r}_{\ell,h}(j).
\end{equation}
This refinement provides additional robustness without altering the single-step nature of HAVE.

\paragraph{Aggregation and projection.}
We aggregate across heads with adaptive weights $w_{\ell,h}$:
\begin{equation}
\label{eq:ctx-agg}
U^{\mathrm{ctx}}_{t}(j)
= \sum_{\ell,h} w_{\ell,h}\,\hat{r}_{\ell,h}(j),
\end{equation}
then project to the vocabulary and confine support to $\mathcal{R}_t$:
\begin{equation}
\label{eq:proj}
U_t(v)
= \sum_{j\in\mathcal{C}} U^{\mathrm{ctx}}_{t}(j)\,\mathbf{1}[\mathrm{tok}(x_j)=v],
\qquad v\in V,
\end{equation}
\begin{equation}
\label{eq:topR}
U_t(v)
\;\leftarrow\;
\frac{U_t(v)\,\mathbf{1}[v\in \mathcal{R}_t]}{\sum_{u\in \mathcal{R}_t} U_t(u)+\varepsilon}.
\end{equation}

\paragraph{Robust fallback.}
To avoid degenerate cases where masked mass vanishes numerically, we revert to a uniform-head average of $\text{Attn}\times V$, ensuring non-zero evidence for fusion (see Sec.~\ref{sec:fusion}).

\subsection{Fusion Policy}
\label{sec:fusion}

The previous sections (\S\ref{sec:hag}--\ref{sec:vc}) define the core evidence constructor $\Phi$: 
HAG produces dynamic head weights $w_{\ell,h}$, and 
VC converts attention and values into a vocabulary-aligned utilization distribution $U_t$ confined to the Top-$R$ candidates. 
We now specify how this utilization signal is integrated with the model distribution to form the final decoding decision.

Given the model distribution $P_t$ and utilization $U_t$, we adopt an uncertainty-scaled \emph{additive residual}:
\begin{equation}
\label{eq:fusion}
S_t \;=\; P_t \;+\; \big(\alpha \cdot H_{\mathrm{norm}}(P_t)\big)\cdot U_t ,
\end{equation}
\begin{equation}
\label{eq:hnorm}
H_{\mathrm{norm}}(P_t)\;=\;\frac{-\sum_{v\in V} P_t(v)\log P_t(v)}{\log |V|}.
\end{equation}
This design amplifies $U_t$ when the model distribution is uncertain, while attenuating it when $P_t$ is already sharp.
Since $U_t$ is restricted to $\mathcal{R}_t$, Eq.~\eqref{eq:fusion} preserves support outside $\mathcal{R}_t$ and is monotone within it (Eqs.~\ref{eq:support}--\ref{eq:monotone}).
If the restricted mass degenerates numerically, the system reverts to a uniform-head $\text{Attn}\!\times\!V$ baseline to guarantee nonzero evidence.

\section{Experiments}
\subsection{Experimental Setup}
\label{sec:exp-setup}

\paragraph{Baselines.} We compare against four representative decoding strategies: (i) \textbf{Greedy} decoding (temperature $0$, top-$p=1$) as a method-agnostic reference; (ii) \textbf{CAD}~\cite{shi2024trusting}; (iii) \textbf{COIECD}~\cite{yuan2024discerning}; and (iv) \textbf{DAGCD}~\cite{huang2025dynamic}. All baselines share the same prompts and retrieval/context construction.

\paragraph{Models.} 
We evaluate HAVE on three widely used open-source LLMs representative of contemporary chat/instruction-tuned architectures: 
\textbf{LLaMA2-7B-Chat}~\cite{touvron2023llama}, \textbf{LLaMA2-13B-Chat}, and \textbf{Mistral-7B-Instruct}~\cite{jiang2023mistral7b}. 
All models are employed \emph{as is}, without finetuning, ensuring that improvements stem solely from decoding-time interventions.

\paragraph{Datasets.} 
We evaluate across five benchmarks covering complementary grounding regimes: 
(i) \textbf{HotpotQA}~\cite{yang2018hotpotqa} (multi-hop reasoning over evidence chains), 
(ii) \textbf{SearchQA}~\cite{dunn2017searchqa} (long-context retrieval with noisy evidence), 
(iii) \textbf{SQuAD}~\cite{rajpurkar2016squad} (single-paragraph reading comprehension), 
(iv) \textbf{Natural Questions (NQ)}~\cite{kwiatkowski2019natural} at the document level, and 
(v) \textbf{NQ-Swap}~\cite{longpre2021entity}, a synthetic variant with simulated conflicts created by swapping or interleaving distractor passages. 
This suite covers both realistic retrieval-augmented settings and controlled stress tests for hallucination robustness.

\paragraph{Evaluation metrics.}
We report standard QA metrics: Exact Match (EM) and token-level F1.

\paragraph{Hyperparameter settings.}
For all experiments, we follow the released DAGCD~\cite{huang2025dynamic} configuration to ensure fairness. Specifically, we adopt the same random seeds, fusion scale $\alpha$, top\_rank, maximum generation length, and dataset sizes as in DAGCD.

We modify the lightweight estimator by using Attn$\times$V features instead of the pure-attention–based variant in DAGCD~\cite{huang2025dynamic}. 

\subsection{Comparative Experiments}

\begin{table}[ht]
\caption{
Performance comparison of different decoding methods across five QA datasets and three LLMs. 
\textbf{Bold} indicates the best performance, and \underline{underlined} the second best. 
Baseline numbers are taken from DAGCD~\cite{huang2025dynamic}, since we follow the same hyperparameter configuration and evaluation protocol for fairness.
}
  \belowrulesep=0pt  
  \aboverulesep=0pt  
  \centering
  \resizebox{\linewidth}{!}{  
    \begin{tabular}{c|c|cc|cc|cc|cc|cc}
    \toprule[1.5pt]
    \multirow{2}{*}{\textbf{LLMs}} & \multirow{2}{*}{\textbf{Decoding}} & \multicolumn{2}{c|}{\textbf{HotpotQA}}  & \multicolumn{2}{c|}{\textbf{SearchQA}} & \multicolumn{2}{c|}{\textbf{SQuAD}} & \multicolumn{2}{c|}{\textbf{NQ}} & \multicolumn{2}{c}{\textbf{NQ-swap}} \\
    \cmidrule{3-12}          &       & \textbf{EM} & \textbf{F1} & \textbf{EM} & \textbf{F1} & \textbf{EM} & \textbf{F1} & \textbf{EM} & \textbf{F1} & \textbf{EM} & \textbf{F1} \\
    \midrule  
    \multirow{4}[0]{*}{\textbf{\makecell{LLaMA2-7B\\-Chat}}} & Greedy & \cellcolor[rgb]{ .949,  .949,  .949}53.33 & \cellcolor[rgb]{ .949,  .949,  .949}67.41  & \cellcolor[rgb]{ .949,  .949,  .949}54.19  & \cellcolor[rgb]{ .949,  .949,  .949}58.11  & \cellcolor[rgb]{ .949,  .949,  .949}67.69  & \cellcolor[rgb]{ .949,  .949,  .949}78.70   & \cellcolor[rgb]{ .949,  .949,  .949}50.47  & \cellcolor[rgb]{ .949,  .949,  .949}65.48  & \cellcolor[rgb]{ .949,  .949,  .949}67.98  & \cellcolor[rgb]{ .949,  .949,  .949}68.78   \\
          & CAD (\emph{NAACL2024})   & 52.86  & 67.16  & 54.16  & 58.11  & 65.89  & 77.91  & 48.89  & 65.00  & 68.04 & 68.85  \\
          & COIECD (\emph{ACL2024})  & 53.14  & 67.03  & \textbf{55.04} & \textbf{58.98} & 68.32 & 79.56   & \underline{52.39} & 66.84  & 69.48  & 70.13  \\
          & DAGCD (\emph{ACL2025}) & \underline{55.31} & \underline{68.61} &  \underline{54.25}  & \underline{58.13}  & \underline{68.49} & \underline{79.76} &  51.69  & \underline{66.92} & \underline{69.50} & \underline{70.30}\\
          & \textbf{HAVE(Ours)} & \textbf{55.65} & \textbf{68.81} &  53.83  & 57.74 & \textbf{69.36} & \textbf{80.37} &  \textbf{52.60}  & \textbf{67.74} & \textbf{70.37} & \textbf{71.07} \\
    \midrule 
    \multirow{4}[0]{*}{\textbf{\makecell{LLaMA2-13B\\-Chat}}} & Greedy & \cellcolor[rgb]{ .949,  .949,  .949}55.01  & \cellcolor[rgb]{ .949,  .949,  .949}69.92  & \cellcolor[rgb]{ .949,  .949,  .949}67.08  & \cellcolor[rgb]{ .949,  .949,  .949}\underline{71.96}  & \cellcolor[rgb]{ .949,  .949,  .949}68.26  & \cellcolor[rgb]{ .949,  .949,  .949}79.45   & \cellcolor[rgb]{ .949,  .949,  .949}53.49  & \cellcolor[rgb]{ .949,  .949,  .949}69.18  & \cellcolor[rgb]{ .949,  .949,  .949}60.69  & \cellcolor[rgb]{ .949,  .949,  .949}61.77   \\
          & CAD (\emph{NAACL2024})  & 54.44  & 69.66 & 67.01  & 71.95  & 66.77  & 78.70   & 52.89  & 68.63  & 60.83  & 61.92    \\
          & COIECD (\emph{ACL2024})  & 56.15  & 70.43  & 67.28  & 71.93  & 68.49 & 80.39  & 53.69 & 69.81 & 62.47  & 63.21   \\
          & DAGCD (\emph{ACL2025}) & \underline{57.76} & \underline{71.69}  & \textbf{68.19} & \textbf{72.73} & \underline{69.66} & \underline{80.76}  & \underline{55.36} & \underline{71.31} & \underline{64.03} & \underline{65.20} \\
          & \textbf{HAVE(Ours)} & \textbf{58.23} & \textbf{71.91} &  \underline{67.40}  & 71.75  & \textbf{70.48} & \textbf{81.22} &  \textbf{55.76}  & \textbf{71.37} & \textbf{64.68} & \textbf{65.81} \\
    \midrule 
    \multirow{4}[1]{*}{\textbf{\makecell{Mistral-7B\\-Instruct}}} & Greedy & \cellcolor[rgb]{ .949,  .949,  .949}58.70  & \cellcolor[rgb]{ .949,  .949,  .949}72.18    & \cellcolor[rgb]{ .949,  .949,  .949}44.42  & \cellcolor[rgb]{ .949,  .949,  .949}49.63  & \cellcolor[rgb]{ .949,  .949,  .949}67.28  & \cellcolor[rgb]{ .949,  .949,  .949}79.37  & \cellcolor[rgb]{ .949,  .949,  .949}52.29  & \cellcolor[rgb]{ .949,  .949,  .949}66.93  & \cellcolor[rgb]{ .949,  .949,  .949}66.90  & \cellcolor[rgb]{ .949,  .949,  .949}67.83  \\
          & CAD (\emph{NAACL2024})   & 49.30  & 64.81   & \underline{45.42}  & \underline{50.96}  & 59.97  & 72.92  & 42.63  & 58.49  & 52.04  & 53.75  \\
          & COIECD (\emph{ACL2024}) & 59.74  & 72.59   & 37.09  & 42.66  & \underline{68.45} & \underline{81.03} & 53.54 & 68.72 & \textbf{72.81} & \textbf{73.81} \\
          & DAGCD (\emph{ACL2025}) & \underline{60.55} & \underline{73.49} & \textbf{47.17} & \textbf{52.65} & 68.30 & 80.62   & \underline{54.78} & \underline{69.72} & 71.38  & 72.12   \\
          & \textbf{HAVE(Ours)} & \textbf{61.26} & \textbf{74.06} &  44.85  & 49.94  & \textbf{69.95} & \textbf{82.11} &  \textbf{55.52}  & \textbf{70.63} & \underline{72.79} & \underline{73.62}\\
    \bottomrule[1.5pt]
    \end{tabular}
}
   \label{tab:results}
   \vspace{-2mm}
\end{table}
Our method (HAVE) consistently outperforms strong baselines across most datasets and models, achieving improvements on 11 out of 15 model–dataset pairs. The gains are most pronounced on reasoning-intensive datasets such as HotpotQA and document-level NQ, while performance on SearchQA remains competitive though slightly below DAGCD.

On \textbf{HotpotQA} and \textbf{NQ}, HAVE yields the largest gains, showing that head-adaptive gating and value calibration are particularly effective for reasoning over long and diverse contexts. \textbf{SQuAD} also benefits, indicating that our method improves grounding even without retrieval noise. On \textbf{NQ-Swap}, HAVE demonstrates robustness under artificially conflicting evidence: it achieves state-of-the-art results on LLaMA2 models and remains highly competitive on Mistral.

Overall, HAVE delivers consistent, model-agnostic improvements across datasets, establishing new state of the art on HotpotQA, SQuAD, and NQ, while matching or surpassing prior approaches on the remaining benchmarks.

\begin{figure}
  \centering
  \includegraphics[scale=0.15]{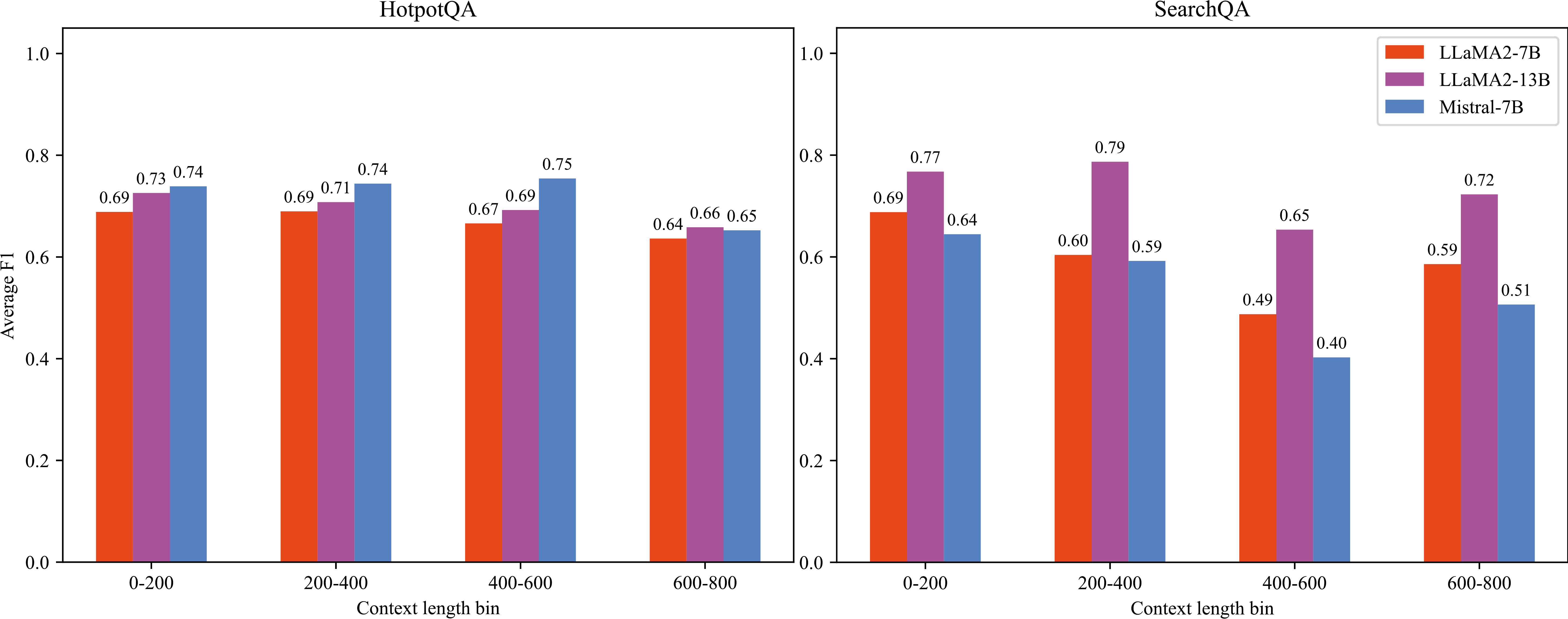}
  \caption{Performance of HAVE across context-length bins on HotpotQA and SearchQA.}\label{fig:data_len}
\end{figure}

\paragraph{By-length analysis.}
Since input length can affect attention patterns during decoding, we further examine HAVE under different context lengths, as shown in Fig.~\ref{fig:data_len}. We partition HotpotQA into four bins (0–200, 200–400, 400–600, 600–800 tokens) and report F1 within each bin. We observe a slight degradation as inputs become longer, which is expected given the increased reasoning difficulty. Nevertheless, the decline is modest, and HAVE remains competitive across all length ranges.

\subsection{Ablation Study}
\label{sec:ablation}

We conduct ablation experiments on the HotpotQA dataset to better understand the contributions of each module and the robustness of hyperparameters in HAVE.

\paragraph{Effectiveness of modules.}
To assess the role of each component, we remove the HAG, the VC, or both modules, and evaluate the performance of HAVE across all three LLMs. 
Table~\ref{tab:ablation-modules} shows that eliminating either HAG or VC consistently leads to performance degradation, while removing both produces the largest drop. 
This indicates that (i) both modules contribute positively and (ii) their effects are complementary: HAG dynamically emphasizes context-relevant heads, whereas VC aligns token-level evidence with write-back contributions. Together, they provide the most significant gains.

\begin{table}[ht]
  \caption{
Ablation study results of the HAVE framework.
}
  \belowrulesep=0pt  
  \aboverulesep=0pt  
  \centering
  \resizebox{0.7\linewidth}{!}{  
    \begin{tabular}{c|cc|cc|cc}
    \toprule[1.5pt]
    \multirow{2}{*}{\textbf{Decoding}} & \multicolumn{2}{c|}{\textbf{LLaMA2-7B-Chat}}  & \multicolumn{2}{c|}{\textbf{LLaMA2-13B-Chat}} & \multicolumn{2}{c}{\textbf{Mistral-7B-Instruct}}\\
    \cmidrule{2-7}  & \textbf{EM} & \textbf{F1} & \textbf{EM} & \textbf{F1} & \textbf{EM} & \textbf{F1}  \\
    \midrule  
    HAVE & 55.65 & 68.81 &  58.23  & 71.91 & 61.26 & 74.06 \\
    w/o HAG & 55.58 & 68.77 &  57.75  & 71.63 & 61.23 & 74.00 \\
    w/o VC & 55.28 & 68.61 &  57.79  & 71.49 & 60.19 & 73.37 \\
    w/o HAG+VC & 54.94 & 68.44 &  57.48  & 71.34 & 60.01 & 73.24 \\
    \bottomrule[1.5pt]
    \end{tabular}
}
   \label{tab:ablation-modules}
   \vspace{-2mm}
\end{table}

\begin{figure}
  \centering
  \includegraphics[scale=0.13]{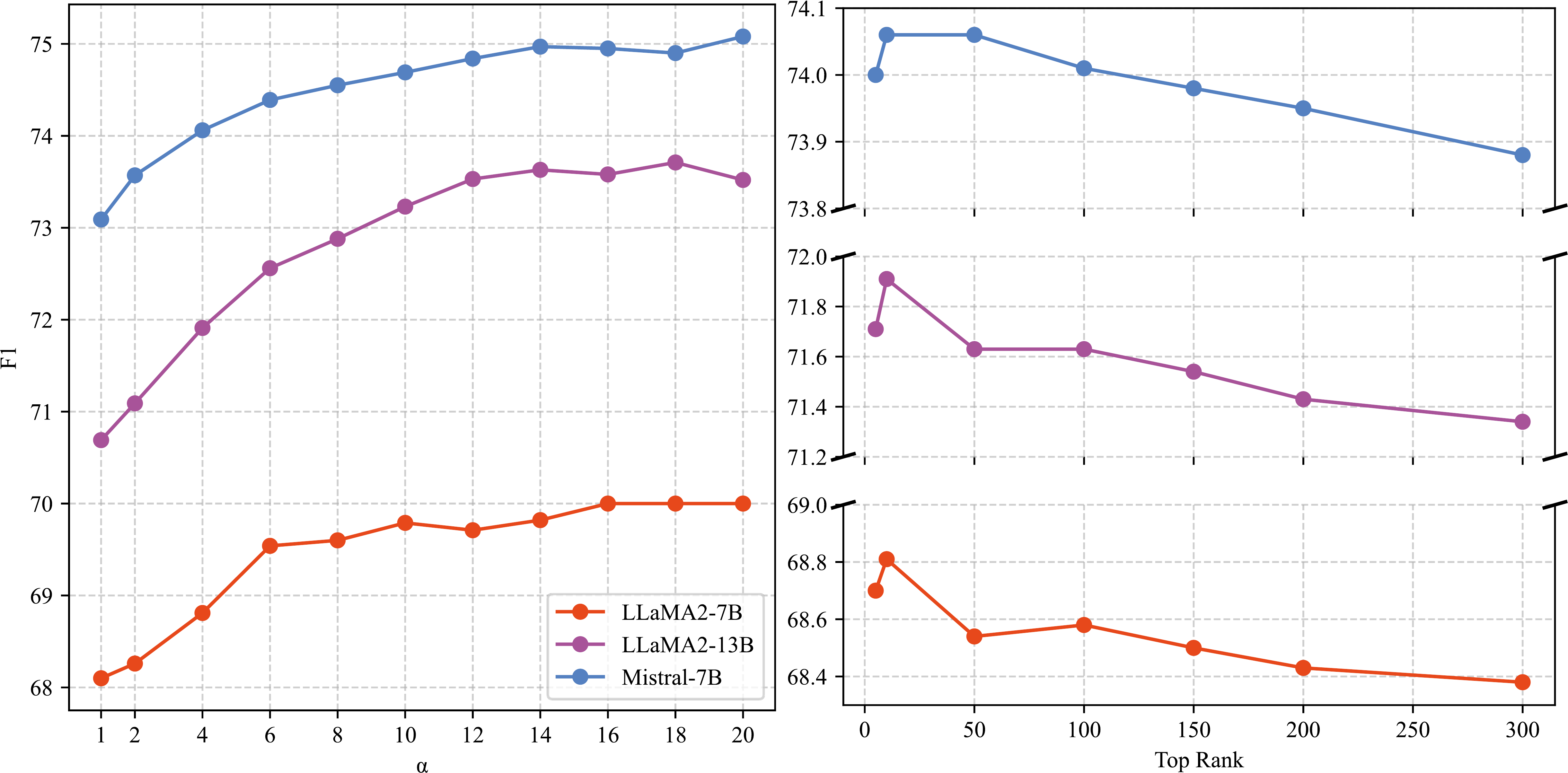}
  \caption{The impact of $\alpha$ and top rank variations on HAVE performance.}\label{fig:alpha_rank}
\end{figure}

\paragraph{Sensitivity to hyperparameters.}
We further examine the sensitivity of HAVE to the fusion scale $\alpha$ and the support size $R$ (\emph{Top-rank}). 
Figure~\ref{fig:alpha_rank} presents the trends: as $\alpha$ increases, performance first improves and then stabilizes, suggesting that a moderate $\alpha$ effectively balances model distribution $P_t$ and evidence $U_t$, while too large values saturate the effect. 
For $R$, increasing the support initially improves results by incorporating more candidate tokens, but overly large $R$ dilutes evidence and causes slight degradation. 
These results demonstrate that HAVE is robust to a wide range of hyperparameters and easy to tune in practice.

\section{Conclusion}
\label{sec:conclusion}

We introduced \textbf{HAVE}, a single-step, parameter-free decoding framework that mitigates hallucination by combining \emph{Head-Adaptive Gating} and \emph{Value Calibration}. 
HAG adaptively reweighs attention heads at the instance level, ensuring that context-relevant heads receive greater emphasis without discarding others. 
VC transforms raw attention into value-augmented token evidence, debiasing sink tokens and better aligning with the model’s write-back contributions. 
Together, these modules construct faithful utilization signals that are fused with the model distribution via an uncertainty-scaled residual, yielding consistent improvements across datasets and model families.

Extensive experiments on five QA datasets demonstrate that HAVE achieves new state-of-the-art performance on multiple benchmarks, with gains that are stable across three open LLMs. 
Ablation studies further confirm that both HAG and VC are indispensable and complementary, while sensitivity analyses show robustness to key hyperparameters.

Beyond empirical gains, HAVE offers practical advantages: it requires no finetuning, introduces only lightweight tensor operations, and is fully compatible with modern attention implementations such as grouped-query attention and sliding-window KV caches. 
These properties make HAVE readily deployable in real-world retrieval-augmented and long-context generation systems.

\paragraph{Future work.} 
An exciting direction is extending HAVE to interactive or multi-turn dialogue~\cite{li2025beyond}, where context relevance shifts dynamically across turns. 
Another avenue is exploring its synergy with retrieval selection or confidence calibration methods. 
We hope HAVE will serve as a transparent, reproducible, and adaptable step toward more trustworthy LLMs.

\section*{Ethical Considerations}
Our work focuses on mitigating hallucinations in LLMs, aiming to improve factuality and reliability. While this can enhance trustworthiness and reduce the spread of misinformation, it does not fully eliminate hallucinations or biases inherited from pretraining data. Downstream users should therefore avoid deploying the method in high-stakes domains (e.g., medical, legal, or safety-critical decision making) without rigorous human oversight. Moreover, as our approach operates at decoding time without altering pretraining corpora, it does not directly address representational biases, fairness, or privacy risks present in LLMs.

\section*{Acknowledgments}
This work was supported by the Graduate Scientific Research and Innovation Project of the People’s Public Security University of China (No.2025yjsky006).

\bibliographystyle{unsrt}  
\bibliography{references}

\begin{thebibliography}{10}

\bibitem{lewis2020retrieval}
Patrick Lewis, Ethan Perez, Aleksandra Piktus, Fabio Petroni, Vladimir Karpukhin, Naman Goyal, Heinrich K{\"u}ttler, Mike Lewis, Wen-tau Yih, Tim Rockt{\"a}schel, et~al.
\newblock Retrieval-augmented generation for knowledge-intensive nlp tasks.
\newblock In {\em Proceedings of the 34th International Conference on Neural Information Processing Systems}, pages 9459--9474, 2020.

\bibitem{huang2025dynamic}
Yanwen Huang, Yong Zhang, Ning Cheng, Zhitao Li, Shaojun Wang, and Jing Xiao.
\newblock Dynamic attention-guided context decoding for mitigating context faithfulness hallucinations in large language models.
\newblock page 5174–5193, 2025.

\bibitem{li2023textbooks}
Yuanzhi Li, S{\'e}bastien Bubeck, Ronen Eldan, Allie Del~Giorno, Suriya Gunasekar, and Yin~Tat Lee.
\newblock Textbooks are all you need ii: phi-1.5 technical report.
\newblock {\em arXiv preprint arXiv:2309.05463}, 2023.

\bibitem{hu2024mitigating}
Minda Hu, Bowei He, Yufei Wang, Liangyou Li, Chen Ma, and Irwin King.
\newblock Mitigating large language model hallucination with faithful finetuning.
\newblock {\em arXiv preprint arXiv:2406.11267}, 2024.

\bibitem{sun2024aligning}
Zhiqing Sun, Sheng Shen, Shengcao Cao, Haotian Liu, Chunyuan Li, Yikang Shen, Chuang Gan, Liangyan Gui, Yu-Xiong Wang, Yiming Yang, et~al.
\newblock Aligning large multimodal models with factually augmented rlhf.
\newblock In {\em Findings of the Association for Computational Linguistics ACL 2024}, pages 13088--13110, 2024.

\bibitem{li2024confidence}
Loka Li, Zhenhao Chen, Guangyi Chen, Yixuan Zhang, Yusheng Su, Eric Xing, and Kun Zhang.
\newblock Confidence matters: Revisiting intrinsic self-correction capabilities of large language models.
\newblock {\em arXiv preprint arXiv:2402.12563}, 2024.

\bibitem{zheng2023can}
Ce~Zheng, Lei Li, Qingxiu Dong, Yuxuan Fan, Zhiyong Wu, Jingjing Xu, and Baobao Chang.
\newblock Can we edit factual knowledge by in-context learning?
\newblock In {\em Proceedings of the 2023 Conference on Empirical Methods in Natural Language Processing}, pages 4862--4876, 2023.

\bibitem{shi2024trusting}
Weijia Shi, Xiaochuang Han, Mike Lewis, Yulia Tsvetkov, Luke Zettlemoyer, and Wen-tau Yih.
\newblock Trusting your evidence: Hallucinate less with context-aware decoding.
\newblock In {\em Proceedings of the 2024 Conference of the North American Chapter of the Association for Computational Linguistics: Human Language Technologies (Volume 2: Short Papers)}, pages 783--791, 2024.

\bibitem{yuan2024discerning}
Xiaowei Yuan, Zhao Yang, Yequan Wang, Shengping Liu, Jun Zhao, and Kang Liu.
\newblock Discerning and resolving knowledge conflicts through adaptive decoding with contextual information-entropy constraint.
\newblock In {\em Findings of the Association for Computational Linguistics ACL 2024}, pages 3903--3922, 2024.

\bibitem{krishna2024genaudit}
Kundan Krishna, Sanjana Ramprasad, Prakhar Gupta, Byron~C Wallace, Zachary~C Lipton, and Jeffrey~P Bigham.
\newblock Genaudit: Fixing factual errors in language model outputs with evidence.
\newblock {\em arXiv preprint arXiv:2402.12566}, 2024.

\bibitem{chen2023converge}
Jiangjie Chen, Rui Xu, Wenxuan Zeng, Changzhi Sun, Lei Li, and Yanghua Xiao.
\newblock Converge to the truth: factual error correction via iterative constrained editing.
\newblock In {\em Proceedings of the Thirty-Seventh AAAI Conference on Artificial Intelligence and Thirty-Fifth Conference on Innovative Applications of Artificial Intelligence and Thirteenth Symposium on Educational Advances in Artificial Intelligence}, pages 12616--12625, 2023.

\bibitem{touvron2023llama}
Hugo Touvron, Louis Martin, Kevin Stone, Peter Albert, Amjad Almahairi, Yasmine Babaei, Nikolay Bashlykov, Soumya Batra, Prajjwal Bhargava, Shruti Bhosale, et~al.
\newblock Llama 2: Open foundation and fine-tuned chat models.
\newblock {\em arXiv preprint arXiv:2307.09288}, 2023.

\bibitem{jiang2023mistral7b}
Albert~Q. Jiang, Alexandre Sablayrolles, Arthur Mensch, Chris Bamford, Devendra~Singh Chaplot, Diego de~las Casas, Florian Bressand, Gianna Lengyel, Guillaume Lample, Lucile Saulnier, Lélio~Renard Lavaud, Marie-Anne Lachaux, Pierre Stock, Teven~Le Scao, Thibaut Lavril, Thomas Wang, Timothée Lacroix, and William~El Sayed.
\newblock Mistral 7b, 2023.

\bibitem{yang2018hotpotqa}
Zhilin Yang, Peng Qi, Saizheng Zhang, Yoshua Bengio, William Cohen, Ruslan Salakhutdinov, and Christopher~D Manning.
\newblock Hotpotqa: A dataset for diverse, explainable multi-hop question answering.
\newblock In {\em Proceedings of the 2018 Conference on Empirical Methods in Natural Language Processing}, pages 2369--2380, 2018.

\bibitem{dunn2017searchqa}
Matthew Dunn, Levent Sagun, Mike Higgins, V~Ugur Guney, Volkan Cirik, and Kyunghyun Cho.
\newblock Searchqa: A new q\&a dataset augmented with context from a search engine.
\newblock {\em arXiv preprint arXiv:1704.05179}, 2017.

\bibitem{rajpurkar2016squad}
Pranav Rajpurkar, Jian Zhang, Konstantin Lopyrev, and Percy Liang.
\newblock Squad: 100,000+ questions for machine comprehension of text.
\newblock In {\em Proceedings of the 2016 Conference on Empirical Methods in Natural Language Processing}, pages 2383--2392, 2016.

\bibitem{kwiatkowski2019natural}
Tom Kwiatkowski, Jennimaria Palomaki, Olivia Redfield, Michael Collins, Ankur Parikh, Chris Alberti, Danielle Epstein, Illia Polosukhin, Jacob Devlin, Kenton Lee, et~al.
\newblock Natural questions: a benchmark for question answering research.
\newblock {\em Transactions of the Association for Computational Linguistics}, 7:453--466, 2019.

\bibitem{longpre2021entity}
Shayne Longpre, Kartik Perisetla, Anthony Chen, Nikhil Ramesh, Chris DuBois, and Sameer Singh.
\newblock Entity-based knowledge conflicts in question answering.
\newblock In {\em Proceedings of the 2021 Conference on Empirical Methods in Natural Language Processing}, pages 7052--7063, 2021.

\bibitem{li2025beyond}
Yubo Li, Xiaobin Shen, Xinyu Yao, Xueying Ding, Yidi Miao, Ramayya Krishnan, and Rema Padman.
\newblock Beyond single-turn: A survey on multi-turn interactions with large language models.
\newblock {\em arXiv preprint arXiv:2504.04717}, 2025.

\end{thebibliography}

\end{document}